\documentclass{article}

\usepackage{microtype}
\usepackage{graphicx}
\usepackage{subfigure}
\usepackage{subcaption}
\usepackage{booktabs} 

\usepackage{hyperref}

\usepackage[accepted]{icml2024}

\usepackage{amsmath}
\usepackage{amssymb}
\usepackage{mathtools}
\usepackage{amsthm}
\usepackage{bm}

\usepackage{amsmath,amsfonts,bm}

\def\eqref#1{equation~\ref{#1}}

\def\1{\bm{1}}

\newcommand{\test}{\mathcal{D_{\mathrm{test}}}}

\DeclareMathAlphabet{\mathsfit}{\encodingdefault}{\sfdefault}{m}{sl}
\SetMathAlphabet{\mathsfit}{bold}{\encodingdefault}{\sfdefault}{bx}{n}

\DeclareMathOperator*{\argmax}{arg\,max}

\usepackage[ruled,vlined]{algorithm2e}
\usepackage{amsfonts}
\usepackage{commath}
\usepackage{svg}

\usepackage[capitalize,noabbrev]{cleveref}

\theoremstyle{plain}
\newtheorem{theorem}{Theorem}[section]

\theoremstyle{definition}

\theoremstyle{remark}

\usepackage[textsize=tiny]{todonotes}

\icmltitlerunning{Accelerating Convergence in Bayesian Few-Shot Classification}

\begin{document}

\twocolumn[
\icmltitle{Accelerating Convergence in Bayesian Few-Shot Classification}



\icmlsetsymbol{equal}{*}

\begin{icmlauthorlist}
\icmlauthor{Tianjun Ke}{yyy}
\icmlauthor{Haoqun Cao}{yyy}
\icmlauthor{Feng Zhou}{yyy,zzz}
\end{icmlauthorlist}

\icmlaffiliation{yyy}{Center for Applied Statistics and School of Statistics, Renmin University of China, Beijing, China}
\icmlaffiliation{zzz}{Beijing Advanced Innovation Center for Future Blockchain and Privacy Computing}

\icmlcorrespondingauthor{Feng Zhou}{feng.zhou@ruc.edu.cn}

\icmlkeywords{Bayesian few-shot classification, mirror descent, Gaussian process classification, variational inference}

\vskip 0.3in]



\printAffiliationsAndNotice{}  

\begin{abstract}
Bayesian few-shot classification has been a focal point in the field of few-shot learning. This paper seamlessly integrates mirror descent-based variational inference into Gaussian process-based few-shot classification, addressing the challenge of non-conjugate inference. By leveraging non-Euclidean geometry, mirror descent achieves accelerated convergence by providing the steepest descent direction along the corresponding manifold. It also exhibits the parameterization invariance property concerning the variational distribution. Experimental results demonstrate competitive classification accuracy, improved uncertainty quantification, and faster convergence compared to baseline models. Additionally, we investigate the impact of hyperparameters and components. 
Code is publicly available at \url{https://github.com/keanson/MD-BSFC}.
\end{abstract}
\vspace{-10pt}

\section{Introduction}
Humans have the ability to learn new skills and adapt to new environments based on very few instances. In contrast, most machine learning techniques, especially deep learning, require vast amounts of examples to achieve similar performance and may yet struggle to generalize. The reason humans are more advanced than machines in adapting and generalizing is that they can leverage prior experience to solve new tasks. It is thus of great interest to design machine learning algorithms that can generalize to novel tasks with a limited amount of data. 

Few-shot classification focuses on the classification of new data given only a limited number of training samples with class labels. It proves particularly useful when collecting training examples is challenging or when annotating data incurs high costs. The scarcity of labeled data in such situations introduces uncertainty over model parameters, commonly referred to as epistemic uncertainty. Effectively managing epistemic uncertainty serves to regularize the model, mitigating the risk of overfitting. Additionally, this uncertainty plays a crucial role in assessing confidence, especially in risk-averse applications like medical diagnosis~\citep{prabhu2019few} and autonomous driving~\citep{bojarski2016end}. 

The Bayesian framework offers a natural approach to capture epistemic uncertainty by introducing a prior distribution over model parameters and computing the posterior using Bayes' theorem based on observed data. In recent years, there has been a significant surge in research applying Bayesian approaches to few-shot learning~\citep{yoon2018bayesian,finn2018probabilistic,ravi2018amortized}. Leveraging the advantages of the Bayesian framework, some recent studies have employed Gaussian processes (GPs) in the context of few-shot classification (FSC), demonstrating competitive performance in accuracy and uncertainty quantification~\citep{massimi2020bayesian,jake2021bayesian, ke2023revisiting}. 

Bayesian inference poses challenges for Gaussian Process (GP) classification due to the non-Gaussian likelihood being non-conjugate to the GP prior, rendering exact posterior computation infeasible. To address this, \citet{massimi2020bayesian} assumes a Gaussian likelihood for class labels, achieving conjugacy in inference. However, this approach is not entirely reasonable since class labels are discrete rather than continuous. \citet{jake2021bayesian, ke2023revisiting} combine P\'{o}lya-Gamma augmentation~\citep{polson2013bayesian} with the one-vs-each softmax approximation~\citep{titsias2016onevseach} and logistic-softmax likelihood separately to establish conditionally conjugate inference. These methods provide improved uncertainty quantification but necessitate the introduction of additional auxiliary variables. 


Differing from the approaches mentioned earlier, we incorporate mirror descent~\citep{nemirovskij1983problem}-based variational inference~\citep{blei2017variational} into GP-based FSC without introducing any auxiliary variables. This approach is consequently named \emph{Mirror Descent based Bayesian Few-Shot Classification} (MD-BFSC). In our method, the optimization within variational inference is accomplished through conjugate computation. Notably, mirror descent leverages non-Euclidean geometry, providing the steepest descent direction along the corresponding manifold, thereby enhancing the convergence rate. It also exhibits the parameterization invariance property concerning the variational distribution~\citep{raskutti2015information}.

Specifically, we make several contributions: \textbf{(1)} We introduce variational inference based on mirror descent to GP-based FSC, thereby transforming non-conjugate inference into an optimization problem with conjugate computation. \textbf{(2)} As demonstrated later, MD-BFSC provides the steepest descent direction along the corresponding non-Euclidean manifold, enhancing the convergence rate and maintaining invariance to the parameterization of the variational distribution. \textbf{(3)} We showcase that our approach achieves competitive classification accuracy and uncertainty quantification, demonstrating a faster convergence rate on standard FSC benchmarks in comparison to few-shot baseline models.

\section{Background}
Our approach involves concepts in few-shot classification, GP classification, variational inference, exponential family, natural gradient descent and mirror descent. We illustrate the basic concepts in each part. The same notation in different sections refers to the same variable unless specified. 

\subsection{Few-shot Classification}
In FSC~\citep{jake2017prototypical,miller2000learning,wang2020generalizing}, a classifier must adapt to new classes which are not observed in training, given only a few samples of each class. Consider a $L$-shot and $C$-way FSC task $s$ with support set $\mathcal{S}_s=\{\mathbf{x}_l,y_l\}_{l=1}^L$ and query set $\mathcal{Q}_s=\{\mathbf{x}_m,y_m\}_{m=1}^M$, $\mathbf{x}_m$ is the input feature and $y_m\in\{1,\ldots,C\}$ is the class label. We can sample distinct tasks from a distribution over tasks to form the training dataset $\mathcal{D_{\text{train}}}=\{\mathcal{S}_s,\mathcal{Q}_s\}_{s=1}^S$. Similarly, the validation dataset $\mathcal{D_{\text{validation}}}$ and test dataset $\mathcal{D_{\text{test}}}$. Given a new task $s^*$ from the test dataset with support and query set $\{\mathcal{S}_{s^*},\mathcal{Q}_{s^*}\}$, our goal is to train a classifier on $\mathcal{S}_{s^*}$ based on the regularity extracted from $\mathcal{D_{\text{train}}}$ to predict the label of samples in $\mathcal{Q}_{s^*}$. The validation dataset is used for tuning hyperparameters, e.g., learning rate. 

\subsection{Gaussian Process Classification with Variational Inference}
Consider a multi-class classification task consisting of $N$ samples with the input features $\mathbf{X}=[\mathbf{x}_1,\ldots,\mathbf{x}_N]^\top$ and the corresponding class labels $\mathbf{y}=[\mathbf{y}_1^\top,\ldots,\mathbf{y}_N^\top]^\top$, where $\mathbf{x}_n\in\mathcal{X}\subset\mathbb{R}^D$ and $\mathbf{y}_n$ is the one-hot encoding for $C$ classes. The multi-class GP classification model~\citep{williams2006gaussian} includes latent GP functions for all classes, i.e., $\{f^1,\ldots,f^C\}$ where $f^c(\cdot):\mathcal{X}\rightarrow\mathbb{R}$ is the latent function for $c$-th class. A GP prior is applied over each latent function $f^c\sim\mathcal{GP}(0,k_{\bm{\eta}^c})$ where $k_{\bm{\eta}^c}$ is the GP kernel for $c$-th class with hyperparameters $\bm{\eta}^c$. Following tradition, we use the softmax likelihood $p(\mathbf{y}_n\vert\mathbf{f}_n)=\prod_{c}\exp{(f_n^c\cdot y_n^c)}/\sum_{c}\exp{(f_n^c)}$ with $f_n^c=f^c(\mathbf{x}_n)$ and $\mathbf{f}_n=[f_n^1,\ldots,f_n^C]^{\top}$, 
and the posterior of latent functions can be computed as: 
\begin{equation}
p(\mathbf{f}\vert\mathbf{y})=\frac{p(\mathbf{y}\vert\mathbf{f})p(\mathbf{f})}{p(\mathbf{y})}=\frac{\prod_{n}p(\mathbf{y}_n\vert\mathbf{f}_n)\prod_{c}p(\mathbf{f}^c)}{\int\prod_{n}p(\mathbf{y}_n\vert\mathbf{f}_n)\prod_{c}p(\mathbf{f}^c)\dif\mathbf{f}},
\label{eq1}
\end{equation}
where $\mathbf{f}^c=[f_1^c,\ldots,f_N^c]^\top$, $\mathbf{f}=[\mathbf{f}^{1\top},\ldots,\mathbf{f}^{C\top}]^\top$, $p(\mathbf{f}^c)=\mathcal{N}(\mathbf{f}^c\vert\mathbf{0},\mathbf{K}^c)$ with $K_{ij}^c=k_{\bm{\eta}^c}(\mathbf{x}_i,\mathbf{x}_j)$. 


However, the exact computation of posterior is infeasible since the non-Gaussian likelihood is non-conjugate to the Gaussian prior. Variational inference (VI)~\citep{blei2017variational,hoffman2013stochastic} is an approximate inference method in which the exact posterior is approximated by a variational distribution $q(\mathbf{f}\vert\bm{\theta})$ with $\bm{\theta}$ being the variational parameter. The optimal variational distribution is obtained by minimising the Kullback-Leibler (KL) divergence between $q$ and the exact posterior, or equivalently maximizing the evidence lower bound (ELBO) $\mathcal{L}(\bm{\theta}):\bm{\Theta}\rightarrow\mathbb{R}$~\citep{bishop2006pattern}: 
\begin{equation*}
\begin{aligned}
\argmax_{\bm{\theta}\in\bm{\Theta}}\mathcal{L}(\bm{\theta})=
\mathbb{E}_{q}\left[\log p(\mathbf{y},\mathbf{f})-\log q(\mathbf{f}\vert\bm{\theta})\right],
\end{aligned}
\label{eq2}
\end{equation*}
where $\bm{\Theta}$ is the set of valid variational parameters. 
In GP classification, the variational distribution $q$ is assumed to be a Gaussian distribution. 
The GP kernel hyperparameter $\bm{\eta}$ can be determined through empirical Bayes by maximizing the marginal likelihood.

\subsection{Exponential Family}
An exponential family~\citep{wainwright2008graphical} is a set of distributions whose probability density function can be expressed in the form:
\begin{equation*}
\begin{aligned}
q(\mathbf{f}\vert\bm{\theta})=h(\mathbf{f})\exp(\bm{\theta}^{\top}\phi(\mathbf{f})-A(\bm{\theta})),
\end{aligned}
\label{eq3}
\end{equation*}
where $\bm{\theta}\in\bm{\Theta}\subset\mathbb{R}^P$ is a vector of natural parameters, $\phi(\mathbf{f})$ is a vector of sufficient statistics, $A(\bm{\theta})=\log\int h(\mathbf{f})\exp(\bm{\theta}^\top \phi(\mathbf{f}))d\mathbf{f}$ is the log-partition function that is convex. An exponential family is referred to as minimal if each distribution has a unique natural parameter. A minimal exponential family distribution can also be parameterized by the mean parameter that is the mean of sufficient statistics: $\bm{\mu}=\mathbb{E}_q[\phi(\mathbf{f})]\in\mathcal{M}\subset\mathbb{R}^P$. The mean parameter also can be obtained by the gradient of log-partition function: $\bm{\mu}=\nabla A(\bm{\theta})$. The convex conjugate function~\citep{rockafellar2015convex} of $A(\bm{\theta})$ is the negative entropy function $H(\bm{\mu})=\mathbb{E}_q[\log q(\mathbf{f}\vert\bm{\theta})]=\bm{\theta}^\top\bm{\mu}-A(\bm{\theta})+\text{const}$. Similarly, the natural parameter can be obtained by the gradient of negative entropy function $\bm{\theta}=\nabla H(\bm{\mu})$. The pair of $\nabla A(\cdot)$ and $\nabla H(\cdot)$ are inverse functions. 

\subsection{Variational Inference with Natural Gradient Descent}
The variational inference using gradient descent can be generalized to that using natural gradient descent~\citep{amari1997neural} which takes the non-Euclidean geometry into account, because each set of parameters corresponds to a distribution~\citep{salimans2013fixed,hoffman2013stochastic,hensman2013gaussian,salimbeni2018natural}. Natural gradient descent assumes the ELBO is maximized over a distribution space and the variational parameters lie on a Riemannian manifold. Given an exponential-family variational distribution $q(\mathbf{f}|\bm{\theta})$, we can denote a Riemannian metric (Fisher information matrix) $\mathbf{I}(\bm{\theta})=-\mathbb{E}_{q}[\nabla_{\bm{\theta}}^2\log q(\mathbf{f}\vert\bm{\theta})]=\nabla^2A(\bm{\theta})$~\citep{rissanen1996fisher,hoffman2013stochastic} to induce a $P$-dimensional Riemannian manifold $(\bm{\Theta},\mathbf{I}(\bm{\theta}))$. To maximize the ELBO on the Riemannian manifold, we can use the following natural gradient update: 
\begin{equation}
\begin{aligned}
\bm{\theta}_{t+1}=\bm{\theta}_t+\rho_t\mathbf{I}^{-1}(\bm{\theta}_t)\nabla\mathcal{L}(\bm{\theta}_t),
\end{aligned}
\label{eq4}
\end{equation}
where $\rho_t$ is the step size. The natural gradient selects the steepest descent direction along the Riemannian manifold and the optimization path on the Riemannian manifold with infinitesimally small steps is invariant to the parameterization of variational distribution~\citep{james2020natural}.

\subsection{Variational Inference with Mirror Descent}
Mirror descent~\citep{nemirovskij1983problem} is another generalization of gradient descent. Given a specific parameterized ELBO $\mathcal{L}(\bm{\mu}):\mathcal{M}\rightarrow\mathbb{R}$ that needs to be maximized, the gradient descent update is $\bm{\mu}_{t+1}=\bm{\mu}_{t}+\rho_t\nabla\mathcal{L}(\bm{\mu}_t)$ where $\rho_t$ is the step size. The above update is equivalent to maximize a local quadratic approximation of $\mathcal{L}(\bm{\mu})$: $\bm{\mu}_{t+1}=\argmax_{\bm{\mu}\in\mathcal{M}}\nabla \mathcal{L}(\bm{\mu}_t)^\top\bm{\mu}-\frac{1}{2\rho_t}\Vert\bm{\mu}-\bm{\mu}_t\Vert_2^2$. Mirror descent replaces Euclidean norm with a proximity function $\Psi(\cdot,\cdot):\mathcal{M}\times\mathcal{M}\rightarrow\mathbb{R}^+$:
\begin{equation*}
\begin{aligned}
\bm{\mu}_{t+1}=\argmax_{\bm{\mu}\in\mathcal{M}}\nabla \mathcal{L}(\bm{\mu}_t)^\top\bm{\mu}-\frac{1}{\rho_t}\Psi(\bm{\mu},\bm{\mu}_t).
\end{aligned}
\label{eq5}
\end{equation*}
The proximity function characterizes the non-Euclidean geometry and different choices of $\Psi$ correspond to different manifolds that variational parameters lie on. 


\section{Methodology}
\begin{figure*}[ht]
    \centering
    \hspace{-30pt}
    \includegraphics[width = 0.73\textwidth]{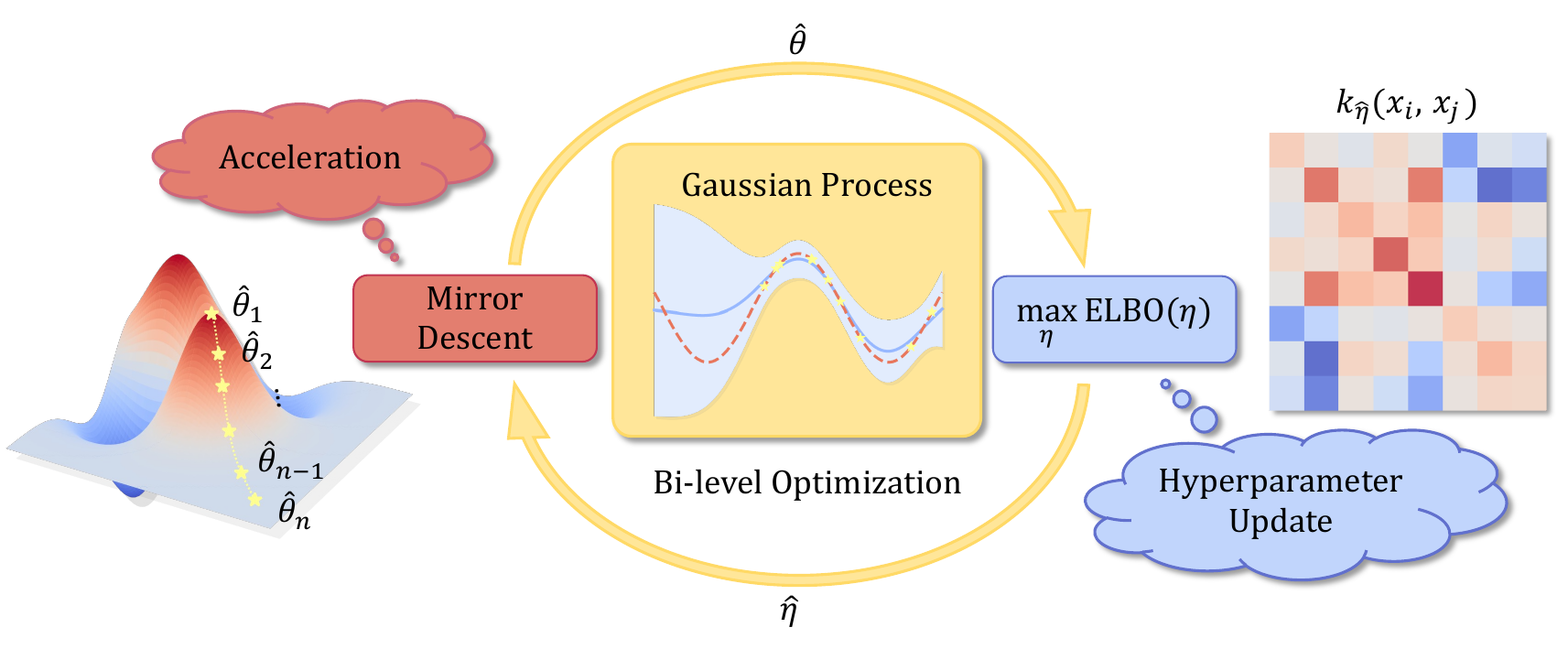}
    \caption{The overview of the training process of MD-BSFC. The diagram illustrates the bi-level optimization process involving an iterative application of mirror descent for VI and hyperparameter tuning.}
    \label{fig:main}
\end{figure*}
The GP-based FSC~\citep{massimi2020bayesian,titsias2020information,jake2021bayesian} is an emerging topic in recent years, which combines the Bayesian framework with few-shot learning. Given a large number of small but related classification tasks, the main idea of GP-based FSC is to perform GP classification for each task and learn a common GP prior that represents the meta-knowledge. This prior is then used to train a GP classifier on a small support set to predict class labels of a query set for a new unseen task. A bi-level optimization framework, which is similar to model-agnostic meta-learning (MAML)~\citep{finn2017model}, is used to learn the common GP prior~\citep{massimi2020bayesian,jake2021bayesian}: in the inner loop, VI is performed to estimate the posterior of latent functions for each task (task-specific parameters); in the outer loop, the GP kernel is designed as a deep kernel~\citep{wilson2016deep} whose hyperparameters (task-common parameters), including the kernel hyperparameters and neural network weights, are updated by empirical Bayes~\citep{maritz2018empirical}. 


In the bi-level optimization framework, both the inner and outer loops consistently employ first-order optimization methods, leading to sluggish convergence in the search for task-common parameters. Moreover, the convergence rate of the inner loop is dependent on the parameterization of the variational distribution, and an arbitrary parameterization may impede faster convergence. 
In this study, our objective is to enhance the convergence rate of the inner loop. Importantly, we do not anticipate the convergence rate of the inner loop to be dependent on the parameterization of the variational distribution since in practice we often lack knowledge about which parameterization is superior. Consequently, the accelerated convergence of the inner loop also facilitates improvements in the convergence of the outer loop. 


\subsection{From Natural Gradient Descent to Mirror Descent}
To address the above issues, an intuitive approach is to optimize the ELBO with natural gradient descent~\citep{amari1997neural} which is a second-order optimization method and possesses the parameterization invariance property in the meanwhile. A large number of existing work applied natural gradient descent to the variational inference of GP~\citep{hensman2013gaussian,malago2015information,salimbeni2018natural}. However, a serious disadvantage hindering the application of natural gradient descent is the complex computation due to the requirement of second information (Fisher information matrix). To facilitate the implementation of natural gradient descent, we propose to use mirror descent in place of natural gradient descent. The following theorem establishes the equivalence between natural gradient descent and mirror descent. A detailed proof is provided in \cref{app2}. 

\begin{theorem}[\citealt{raskutti2015information}]
Given two parameterized ELBO with mean parameter and natural parameter $\widetilde{\mathcal{L}}(\bm{\mu})=\mathcal{L}(\bm{\theta})$, to maximize the ELBO, the mirror descent over the mean parameter
\begin{equation}
\begin{aligned}
\bm{\mu}_{t+1}=\argmax_{\bm{\mu}\in\mathcal{M}}\nabla \widetilde{\mathcal{L}}(\bm{\mu}_t)^\top\bm{\mu}-\frac{1}{\rho_t}B_{H}(\bm{\mu},\bm{\mu}_t),
\end{aligned}
\label{eq6}
\end{equation}
using the Bregman divergence~\citep{bregman1967relaxation} $B_H(\bm{\mu},\bm{\mu}_t)=H(\bm{\mu})-H(\bm{\mu}_t)-\nabla H(\bm{\mu}_t)^\top(\bm{\mu}-\bm{\mu}_t)$ induced by the negative entropy function $H(\cdot)$ is equivalent to the natural gradient descent over the natural parameter
\begin{equation*}
\begin{aligned}
\bm{\theta}_{t+1}=\bm{\theta}_{t}+\rho_t[\nabla^2_{\bm{\theta}}A(\bm{\theta}_t)]^{-1}\nabla_{\bm{\theta}}\mathcal{L}(\bm{\theta}_t),
\end{aligned}
\label{eq7}
\end{equation*}
using the Riemannian metric induced by the log-partition function $A(\cdot)$. 
\label{theorem1}
\end{theorem}

The theorem above asserts that employing natural gradient descent over the natural parameter using Fisher information, induced by the log-partition function, is equivalent to implementing mirror descent over the mean parameter using Bregman divergence induced by the negative entropy function. The notable advantage of mirror descent over natural gradient descent lies in its reliance on only the first-order gradient, as opposed to the second-order gradient required by the latter. 
Consequently, the implementation of natural gradient descent can be reformulated as a computationally more efficient mirror descent. 


\subsection{From Mirror Descent to Conjugate Bayesian Inference}
As demonstrated by \citet{khan2017conjugate}, \cref{eq6} can be additionally streamlined into a Bayesian inference within a conjugate model. The subsequent theorem establishes the equivalence between \cref{eq6} and Bayesian inference within a conjugate model. We refer to \citet{khan2017conjugate} for the complete proof. 

\begin{theorem}[\citealt{khan2017conjugate}]
The mirror descent in \cref{eq6} is equivalent to the following conjugate Bayesian inference: 
\begin{gather} \nonumber
q(\mathbf{f}\vert\bm{\theta}_{t+1})\propto\exp{(\widetilde{\bm{\theta}}_t^\top \phi(\mathbf{f}))}p(\mathbf{f}\vert\bm{\eta}),\label{eq8}\\
\widetilde{\bm{\theta}}_t=(1-\rho_t)\widetilde{\bm{\theta}}_{t-1}+\rho_t\nabla_{\bm{\mu}}\mathbb{E}_q[\log p(\mathbf{y}\vert\mathbf{f})]\vert_{\bm{\mu}=\bm{\mu}_t},
\label{eq9}
\end{gather}
with $\widetilde{\bm{\theta}}_0=\mathbf{0}$ and $\bm{\theta}_1=\bm{\eta}$. Because $p(\mathbf{f}\vert\bm{\eta})$ is the Gaussian prior in \cref{eq1} that is also a exponential-family distribution, $\bm{\theta}_{t+1}$ can be obtained by conjugate computation $\bm{\theta}_{t+1}=\widetilde{\bm{\theta}}_t+\bm{\eta}$. 
\label{theorem2}
\end{theorem}
The aforementioned theorem asserts that the original softmax likelihood in \cref{eq1} is approximated by a Gaussian distribution, with its natural parameter iteratively updated by the gradient of the expectation of the log-likelihood. Subsequently, the variational parameter is updated by adding the natural parameters of the approximated Gaussian distribution and the prior. 

The motivation for using mirror descent should now be clear: for the current problem, mirror descent over the mean parameter is an equivalent approach to natural gradient descent over the natural parameter. Therefore, it also exhibits a second-order convergence rate and possesses the parameterization invariance property. Moreover, mirror descent streamlines the optimization process by requiring only first-order information, and the optimization step can be simplified as a conjugate Bayesian inference computation. 

\subsection{Algorithm}
The bi-level optimization framework of MD-BFSC is depicted in \cref{fig:main} and summarized as follows. In the inner loop, the variational parameters $\bm{\theta}$ (task-specific parameters) are updated by using \cref{eq8,eq9}. For $\nabla_{\bm{\mu}}\mathbb{E}_q[\log p(\mathbf{y}\vert\mathbf{f})]$ in \cref{eq9}, we assume the variaitonal distribution $q(\mathbf{f}\vert\bm{\theta})=\prod_c q(\mathbf{f}^c\vert\bm{\theta}^c)$. Since $p(\mathbf{y}\vert\mathbf{f})=\prod_n p(\mathbf{y}_n\vert\mathbf{f}_n)$, we can compute the gradient of each point $\nabla_{\bm{\mu}_n}\mathbb{E}_{q(\mathbf{f}_n)}[\log p(\mathbf{y}_n\vert\mathbf{f}_n)]$ separately. If we define $\mathbf{m}_n$ and $\mathbf{v}_n$ to be the mean and covariance diagonal (non-diagonal entries are $0$) of the marginal distribution $q(\mathbf{f}_n)$,
then $\nabla_{\bm{\mu}_n}\mathbb{E}_{q(\mathbf{f}_n)}[\log p(\mathbf{y}_n\vert\mathbf{f}_n)]$ can be expressed as: 
\begin{gather*}
\nabla_{\bm{\mu}_n^{(1)}}\mathbb{E}_{q(\mathbf{f}_n)}[\log p(\mathbf{y}_n\vert\mathbf{f}_n)]=\mathbf{g}_{\mathbf{m}_n}-2\mathbf{g}_{\mathbf{v}_n} \odot \mathbf{m}_n,\label{eq10}\\
\nabla_{\bm{\mu}_n^{(2)}}\mathbb{E}_{q(\mathbf{f}_n)}[\log p(\mathbf{y}_n\vert\mathbf{f}_n)]=\mathbf{g}_{\mathbf{v}_n},\label{eq11}
\end{gather*}
where $\bm{\mu}_n^{(1)}$ and $\bm{\mu}_n^{(2)}$ are the two mean parameters of $q(\mathbf{f}_n)$, $\mathbf{g}_{\mathbf{m}_n}=\nabla_{\mathbf{m}_n}\mathbb{E}_{q}[\log p(\mathbf{y}_n\vert\mathbf{f}_n)]=\mathbb{E}_{q}\left[\mathbf{y}_n-\frac{\exp{(\mathbf{f}_n)}}{\sum_c \exp{(\mathbf{f}_n)}}\right]$, $\mathbf{g}_{\mathbf{v}_n}=\nabla_{\mathbf{v}_n}\mathbb{E}_{q}[\log p(\mathbf{y}_n\vert\mathbf{f}_n)]=\frac{1}{2}\mathbb{E}_{q}\left[\frac{\exp{(2\mathbf{f}_n)}}{(\sum_c\exp{(\mathbf{f}_n)})^2}-\frac{\exp{(\mathbf{f}_n)}}{\sum_c\exp{\mathbf{f}_n}}\right]$, $\odot$ is the element-wise product~\citep{opper2009variational}. A proof of \cref{eq10,eq11} is provided in \cref{app3}. 

In the outer loop, the GP kernel is designed as a deep kernel~\citep{wilson2016deep} and the kernel hyperparameter $\bm{\eta}$ includes the traditional kernel hyperparameters and neural network weights. The hyperparameter $\bm{\eta}$ (task-common parameters) is updated by maximizing the ELBO, that is an approximation of log-marginal likelihood: 
\begin{equation}
\begin{aligned}
\argmax_{\bm{\eta}}\mathcal{L}(\bm{\eta})=
\mathbb{E}_{q}\left[\log p(\mathbf{y}\vert\mathbf{f})+\log p(\mathbf{f}\vert\bm{\eta})-\log q(\mathbf{f})\right].
\end{aligned}
\label{eq12}
\end{equation}
For prediction in a new test task with support set: $\mathcal{S}=\{\mathbf{X},\mathbf{y}\}$ and query set without labels $\mathcal{Q}=\mathbf{X}_*$, the label predictive distribution of one sample $\mathbf{x}_*$ in the query set is: 
\begin{subequations}
\label{eq15}
\begin{align}
\label{eq15a}
&~ p(y_*=r\mid\mathbf{x}_*,\mathbf{X},\mathbf{y},\hat{\bm{\eta}}) \\ \nonumber = & ~\int p(y_*=r\mid\mathbf{f}_*)\prod_{c=1}^C q(f_*^c\mid\mathbf{X},\mathbf{Y},\hat{\bm{\eta}})d\mathbf{f}_*,\\ 
\label{eq15b} 
&~ 
q(f_*^c\mid\mathbf{X},\mathbf{y},\hat{\bm{\eta}}) \\ \nonumber = & ~ \int p(f_*^c\mid\mathbf{f}^c)q(\mathbf{f}^c\mid\mathbf{X},\mathbf{y},\hat{\bm{\eta}})d\mathbf{f}^c=\mathcal{N}(f_*^c\mid\mu_*^c,{\sigma_*^2}^c),
\end{align}
\end{subequations}
where $\hat{\bm{\eta}}$ is the estimated kernel hyperparameter from the training dataset, $\mathbf{f}_*=[f^1(\mathbf{x}_*),\ldots,f^C(\mathbf{x}_*)]^{\top}$, $q(\mathbf{f}^c\mid\mathbf{X},\mathbf{y},\hat{\bm{\eta}})$ is a Gaussian distribution computed by \cref{eq8,eq9}. 
By converting the natural parameters into traditional parameters, $\mathcal{N}(\mathbf{f}^c\vert\bm{\theta}^c)\to \mathcal{N}(\mathbf{f}^c\vert\mathbf{m}^c,\bm{\Sigma}^c)$, we can derive $\mu_*^c=\mathbf{k}_{*L}^c\mathbf{K}_{LL}^{c^{-1}}\mathbf{m}^c$ and ${\sigma_*^2}^c=k_{**}^c-\mathbf{k}_{*L}^c\mathbf{K}_{LL}^{c^{-1}}\mathbf{k}_{L*}^c+\mathbf{k}_{*L}^c\mathbf{K}_{LL}^{c^{-1}}\bm{\Sigma}^c\mathbf{K}_{LL}^{c^{-1}}\mathbf{k}_{L*}^c$ 
where $L$ is the number of samples in the support set of the test dataset, and all the $\mathbf{k}$'s are the respective kernel matrices. 
The integral in \cref{eq15a} is intractable, necessitating the use of Monte Carlo for computation. The training and test process of MD-BFSC is summarized in \cref{alg1}.

\begin{algorithm}[t]
\SetAlgoLined
\SetKwInput{training}{Training}
\SetKwInput{test}{Test}
\SetKwInput{Input}{Input}
\SetKwInput{Output}{Output}
\begin{center}
\begin{minipage}[t]{0.45\textwidth}
\training{}
\Input{Input feature and class labels for $S$ tasks: $\{\mathbf{X}_s\}_{s=1}^S$, $\{\mathbf{y}_s\}_{s=1}^S$}
\Output{GP kernel hyperparameter $\bm{\eta}$}
Initialize GP kernel hyperparameter $\bm{\eta}$ and variational parameters $\widetilde{\bm{\theta}}_0=\mathbf{0}$ and $\bm{\theta}_1=\bm{\eta}$\;
\For{Iteration}{
\For{Task $s$}{
\# Update task-specific parameters

\For{Step $t$}{ 
  Update $\widetilde{\bm{\theta}}_t^s$ by \cref{eq9,eq10,eq11}\;
  Update $\bm{\theta}_{t+1}^s=\widetilde{\bm{\theta}}_t^s+\bm{\eta}$\;
  }
  \# Update task-common parameters
  
  Update $\bm{\eta}$ by \cref{eq12}. 
  }
 }
\end{minipage}
\hspace{0.09in}
\begin{minipage}[t]{0.45\textwidth}
\test{}
\Input{Support set $\mathcal{S}=\{\mathbf{X},\mathbf{y}\}$; query set $\mathcal{Q}=\mathbf{X}_*$; learned hyperparameter $\hat{\bm{\eta}}$}
\Output{Predicted labels}
Initialize variational parameters $\widetilde{\bm{\theta}}_0=\mathbf{0}$ and $\bm{\theta}_1=\hat{\bm{\eta}}$\;

\# Update task-specific parameters
  
\For{Step $t$}{ 
  Update $\widetilde{\bm{\theta}}_t$ by \cref{eq9,eq10,eq11}\;
  Update $\bm{\theta}_{t+1}=\widetilde{\bm{\theta}}_t+\hat{\bm{\eta}}$\;
  }
\# Predict labels

\For{$\mathbf{x}_*\in\mathbf{X}_*$}{ 
Predict $y_*$ by \cref{eq15}. 
}
\end{minipage}
\end{center}
\caption{Mirror Descent based Bayesian Few-Shot Classification}
\label{alg1}
\end{algorithm}


\section{Experiments}
In this section, we present the performance of MD-BFSC on few-shot classification tasks, encompassing accuracy, uncertainty quantification, and convergence rate. We address three challenging tasks using benchmark datasets, including Caltech-UCSD Birds \citep{wah2011caltech}, mini-ImageNet \citep{ravi2017optimization}, Omniglot \citep{lake2011one}, and EMNIST \citep{cohen2017emnist}.

\begin{table*}[t]
    \centering
    \caption{Accuracy performance for all models in 1-shot and 5-shot 5-way few-shot classification tasks. Baseline results are from \citet{jake2021bayesian} and \citet{ke2023revisiting}. Results are evaluated over 5 batches of 600 episodes with different random seeds.}
    \scalebox{0.9}{
    \begin{tabular}{lcccccc}
         \toprule & \multicolumn{2}{c}{\text { \textbf{CUB} }} & \multicolumn{2}{c}{\text { \textbf{mini-ImageNet} } $\mathbf{\rightarrow}$ \text { \textbf{CUB} }} & \multicolumn{2}{c}{\text { \textbf{Omniglot} } $\mathbf{\rightarrow}$ \text { \textbf{EMNIST} }} \\
        \text { \textbf{Method} } & \text { \textbf{1-shot} } & \text { \textbf{5-shot} } & \text { \textbf{1-shot} } & \text { \textbf{5-shot} } & \text { \textbf{1-shot} } & \text { \textbf{5-shot} } \\
        \midrule \text { \textbf{Feature Transfer} } & 46.19 $\pm$ 0.64 & 68.40 $\pm$ 0.79 & 32.77 $\pm$ 0.35 & 50.34 $\pm$ 0.27 & 64.22 $\pm$ 1.24 & 86.10 $\pm$ 0.84\\
        \text { \textbf{Baseline++} } & 61.75 $\pm$ 0.95 & 78.51 $\pm$ 0.59 & 39.19 $\pm$ 0.12 & \textbf{57.31} $\pm$ \textbf{0.11} & 56.84 $\pm$ 0.91 & 80.01 $\pm$ 0.92 \\
        \text { \textbf{MatchingNet} } & 60.19 $\pm$ 1.02 & 75.11 $\pm$ 0.35 & 36.98 $\pm$ 0.06 & 50.72 $\pm$ 0.36 & 75.01 $\pm$ 2.09 & 87.41 $\pm$ 1.79 \\
        \text { \textbf{ProtoNet} } & 52.52 $\pm$ 1.90 & 75.93 $\pm$ 0.46 & 33.27 $\pm$ 1.09 & 52.16 $\pm$ 0.17 & 72.04 $\pm$ 0.82 & 87.22 $\pm$ 1.01 \\
        \text { \textbf{RelationNet} } & 62.52 $\pm$ 0.34 & 78.22 $\pm$ 0.07 & 37.13 $\pm$ 0.20 & 51.76 $\pm$ 1.48 &  75.62 $\pm$ 1.00 & 87.84 $\pm$ 0.27 \\
        \text { \textbf{MAML} } & 56.11 $\pm$ 0.69 & 74.84 $\pm$ 0.62 & 34.01 $\pm$ 1.25 & 48.83 $\pm$ 0.62 & 72.68 $\pm$ 1.85 & 83.54 $\pm$ 1.79 \\
        \text { \textbf{DKT} } & 63.37 $\pm$ 0.19 & 77.73 $\pm$ 0.26 & 40.22 $\pm$ 0.54 & 55.65 $\pm$ 0.05 & 73.06 $\pm$ 2.36 & \textbf{88.10 $\pm$ 0.78} \\
        \text { \textbf{Bayesian MAML} } & 55.93 $\pm$ 0.71 & 72.87 $\pm$ 0.26 & 33.52 $\pm$ 0.36 & 51.35 $\pm$ 0.16 & 63.94 $\pm$ 0.47 & 65.26 $\pm$ 0.30 \\
        \text { \textbf{Bayesian MAML (Chaser)} } & 53.93 $\pm$ 0.72 & 71.16 $\pm$ 0.32 & 36.22 $\pm$ 0.50 & 51.53 $\pm$ 0.43 & 55.04 $\pm$ 0.34 & 54.19 $\pm$ 0.32 \\
        \text { \textbf{ABML} } & 49.57 $\pm$ 0.42 & 68.94 $\pm$ 0.16 & 29.35 $\pm$ 0.26 & 45.74 $\pm$ 0.33 & 73.89 $\pm$ 0.24 & 87.28 $\pm$ 0.40 \\
        \text { \textbf{OVE PG GP (ML)} } & 63.98 $\pm$ 0.43 & 77.44 $\pm$ 0.18 & 39.66 $\pm$ 0.18 & 55.71 $\pm$ 0.31 & 68.43 $\pm$ 0.67 & 86.22 $\pm$ 0.20 \\
        \text { \textbf{OVE PG GP (PL)}} & 60.11 $\pm$ 0.26 & 79.07 $\pm$ 0.05 & 37.49 $\pm$ 0.11 & 57.23 $\pm$ 0.31 & \textbf{77.00 $\pm$ 0.50} & 87.52 $\pm$ 0.19 \\
        \text { \textbf{CDKT (ML) ($\tau < 1$)} } & 65.21 $\pm$ 0.45 & \textbf{79.10} $\pm$ \textbf{0.33} & 40.43 $\pm$ 0.43 & 55.72 $\pm$ 0.45 & - & - \\
        \text { \textbf{CDKT (ML) ($\tau = 1$)} } & 60.85 $\pm$ 0.38 & 75.98 $\pm$ 0.33 & 35.57 $\pm$ 0.30 & 52.42 $\pm$ 0.50 & - & - \\
        \text { \textbf{CDKT (PL) ($\tau < 1$)} } & 59.49 $\pm$ 0.35 & 76.95 $\pm$ 0.28 & 39.18 $\pm$ 0.34 & 56.18 $\pm$ 0.28 & - & - \\
        \text { \textbf{CDKT (PL) ($\tau = 1$)} } & 52.91 $\pm$ 0.29 & 73.34 $\pm$ 0.40 & 37.62 $\pm$ 0.32 & 54.32 $\pm$ 0.19 & - & - \\
        \midrule
        \text { \textbf{MD-BFSC} } & \textbf{65.45 $\pm$ 0.42} & 78.38 $\pm$ 0.21 & \textbf{40.75 $\pm$ 0.31} & 56.98 $\pm$ 0.30 & 74.02 $\pm$ 0.49 & 87.05 $\pm$ 0.23 \\   
        \bottomrule
    \end{tabular}
    }
    \label{tab:fsc}
\end{table*}

\subsection{Accuracy}
\label{sec:acc}
In this section, we present the experimental arrangement and disclose the findings of our few-shot classification tasks. While we acknowledge the existence of various configurations for few-shot classification, we opt for the basic setup in Bayesian meta-learning for a fair comparison. Consistent with the methodology employed in previous studies \citep{massimi2020bayesian}, we utilize a standard Conv4 architecture \citep{vinyals2016matching} as the underlying backbone, and evaluate our model across six distinct scenarios, encompassing 1-shot and 5-shot situations for both in-domain and cross-domain tasks. We subject our model to a comprehensive assessment against a range of baselines and state-of-the-art models, including Feature Transfer \citep{chen2019fewshot}, Baseline++ \citep{chen2019fewshot}, MatchingNet \citep{vinyals2016matching}, ProtoNet \citep{snell2017prototypical}, RelationNet \citep{sung2018learning}, MAML \citep{finn2017model}, DKT \citep{massimi2020bayesian}, Bayesian MAML \citep{yoon2018bayesian}, ABML \citep{ravi2019amortized}, OVE \citep{jake2021bayesian}, and LS \citep{ke2023revisiting}. It is worth noting that DKT, LS, and OVE share similarities with our proposed approach as they are all GP-based models with different likelihood functions and inference methods. The default number of epochs from \citet{ke2023revisiting} is employed for training MD-BSFC. During training, we run 3 steps for the inner loop updates in all experiments due to its fast convergence and then conduct a 1 step update for the hyperparameters with Adam. For further experimental details, please refer to \cref{sec:exp}. 

We report the average accuracy and standard deviation of our model evaluated on 5 batches of 600 episodes with different random seeds in \cref{tab:fsc}. Our model achieves the highest accuracy in 1-shot experiments on the CUB dataset (65.45\%, in-domain) and the CUB $\mathbf{\rightarrow}$ mini-ImageNet dataset (40.75\%, cross-domain). As for other datasets and settings, our model also achieves near-optimal or comparable results. While the main strength of our method is its theoretically fast convergence, we find that its classification accuracy is comparable to state-of-the-art models, demonstrating its utility beyond rapid learning.

\subsection{Uncertainty Quantification}
In the realm of FSC, the quantification of uncertainty holds great significance due to the potential high uncertainty in predictions made by classifiers trained with limited data. Particularly in high-risk domains, it is crucial for a robust model to effectively characterize such uncertainty. 

To quantify uncertainty, we employ two widely-used metrics: the expected calibration error (ECE) \citep{guo2017calibration} and the maximum calibration error (MCE) \citep{ovadia2019can}. ECE measures the average difference between confidence (probability outputs) and accuracy within predefined bins, while MCE is similar but focuses on the maximum difference. Following the evaluation protocol outlined by \citet{massimi2020bayesian}, we compute the ECE and MCE on the test set. The summarized results of the 5-shot experiments can be found in \cref{tab:ece}. Specifically, we achieve the lowest ECE values on CUB (0.005, in-domain) and mini-ImageNet $\mathbf{\rightarrow}$ CUB (0.005, cross-domain), as well as the lowest MCE values on mini-ImageNet $\mathbf{\rightarrow}$ CUB (0.014, cross-domain) and Omniglot $\mathbf{\rightarrow}$ EMNIST (0.024, cross-domain). While our model slightly lags behind the state-of-the-art models in other scenarios, the overall outcome highlights its remarkable reliability in uncertainty calibration, thereby demonstrating promising robustness in few-shot scenarios. 

\begin{table*}[t]
    \centering
    \caption{ECE and MCE performance for all models in 5-shot 5-way tasks on CUB, mini-ImageNet $\mathbf{\rightarrow}$ CUB, and Omniglot $\mathbf{\rightarrow}$ EMNIST. Baseline results are from \citet{jake2021bayesian} and \citet{ke2023revisiting}. All metrics are computed on 3,000 random tasks from the test set.}
    \scalebox{0.9}{
    \begin{tabular}{lcccccc}
         \toprule & \multicolumn{2}{c}{\text { \textbf{CUB} }} & \multicolumn{2}{c}{\text { \hspace{-12.5pt}\textbf{mini-ImageNet} $\mathbf{\rightarrow}$ \textbf{CUB} }} & \multicolumn{2}{c}{\text {\hspace{-12.5pt} \textbf{Omniglot} } $\mathbf{\rightarrow}$ \text { \textbf{EMNIST} }} \\
        \text { \textbf{Method} } & \text { \textbf{ECE} } & \text { \textbf{MCE} } & \text { \textbf{ECE} } & \text { \textbf{MCE} } & \text { \textbf{ECE} } & \text { \textbf{MCE} } \\
        \midrule 
        \text { \textbf{Feature Transfer} } & 0.187 & 0.250 & 0.275 & 0.646 & \textbf{0.004} & 0.092  \\
        \text { \textbf{Baseline++} } & 0.421 & 0.502 & 0.315 & 0.537 & 0.390 & 0.475  \\
        \text { \textbf{MatchingNet} } & 0.023 & 0.031 & 0.030 & 0.079 & 0.057 & 0.259 \\
        \text { \textbf{ProtoNet} } &  0.034 & 0.059 & 0.009 & 0.025 & 0.010 & 0.243 \\
        \text { \textbf{RelationNet} } &  0.438 & 0.593 & 0.234 & 0.554 & 0.552 & 0.594  \\
        \text { \textbf{DKT} } &  0.187 & 0.250 & 0.236 & 0.426 & 0.413 & 0.483 \\
        \text { \textbf{Bayesian MAML} } &  0.018 & 0.047 & 0.048 & 0.077 & 0.090 & 0.124 \\
        \text { \textbf{Bayesian MAML (Chaser)} } &  0.047 & 0.104 & 0.066 & 0.260 & 0.235 & 0.306  \\
        \text { \textbf{OVE PG GP (ML)} } &  0.026 & 0.043 & 0.049 & 0.066 & 0.112 & 0.183 \\
        \text { \textbf{OVE PG GP (PL)}} &  \textbf{0.005} & \textbf{0.023} & 0.020 & 0.032 & 0.064 & 0.240  \\
        \text { \textbf{CDKT (ML)} } & \textbf{0.005} & 0.036 & 0.007 & 0.020 & - & -  \\
        \text { \textbf{CDKT (PL)}} & 0.018 & 0.223 & 0.010 & 0.029 & - & - \\
        \midrule
        \text { \textbf{MD-BFSC} } & \textbf{0.005} & 0.067 & \textbf{0.005} & \textbf{0.014} & 0.010 & \textbf{0.025}  \\
        \bottomrule
    \end{tabular}}
    \label{tab:ece}
\end{table*}

\begin{figure}[btp]
\centering
    \subfigure[CUB]{
        \includegraphics[width = 0.33\linewidth]{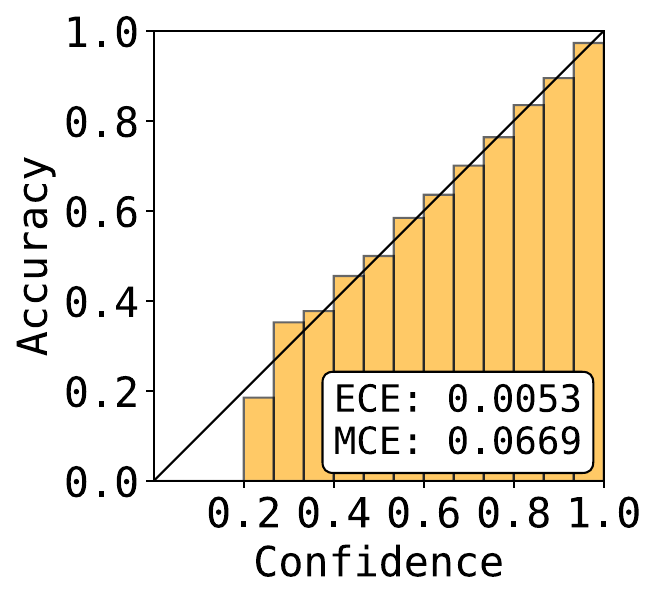}
        \label{fig:CUBELBO}
    }
    \hspace{-15pt}
    \subfigure[MI $\mathbf{\rightarrow}$ CUB]{
        \includegraphics[width = 0.33\linewidth]{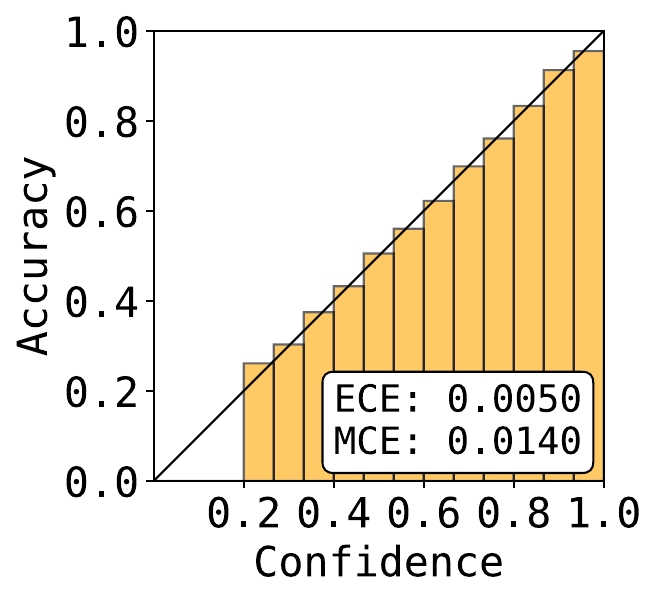}
        \label{fig:MINIELBO}
    }
    \hspace{-15pt}
    \subfigure[Omni $\mathbf{\rightarrow}$ EMNIST]{
        \includegraphics[width = 0.33\linewidth]{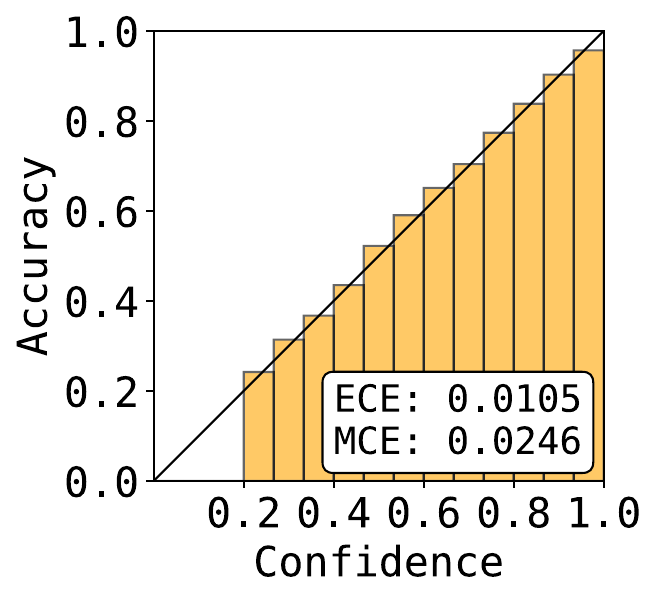}
        \label{fig:CROSSELBO}
    }
    \caption{Reliability diagrams on 5-shot classification with ECE and MCE metrics. MI denotes mini-ImageNet and Omni denotes Omniglot. Results are computed on 3,000 test tasks.}
\label{fig:reliability}
\end{figure}

The reliability diagrams displayed in \cref{fig:reliability} illustrate the performance of our model in terms of reliability. A robust uncertainty quantification model should exhibit a confidence barplot that aligns with the diagonal line. Our model closely adheres to the diagonal line. 

\vspace{-5pt}
\subsection{Convergence Rate}
In this section, we would like to investigate the empirical convergence rate of our proposed method. 
Therefore, we compare MD-BSFC to the method that employs the exact same variational distribution, but its inner-loop VI is implemented through vanilla gradient descent. 
We disentangle the bi-level optimization into two parts: fixing the hyperparameters of the kernel and neural network to check inner-loop convergence and checking the outer-loop convergence with full bi-level optimization. 

\begin{figure*}[ht]
    \centering
    \subfigure[1-shot classification]{
        \includegraphics[width = 0.48\textwidth]{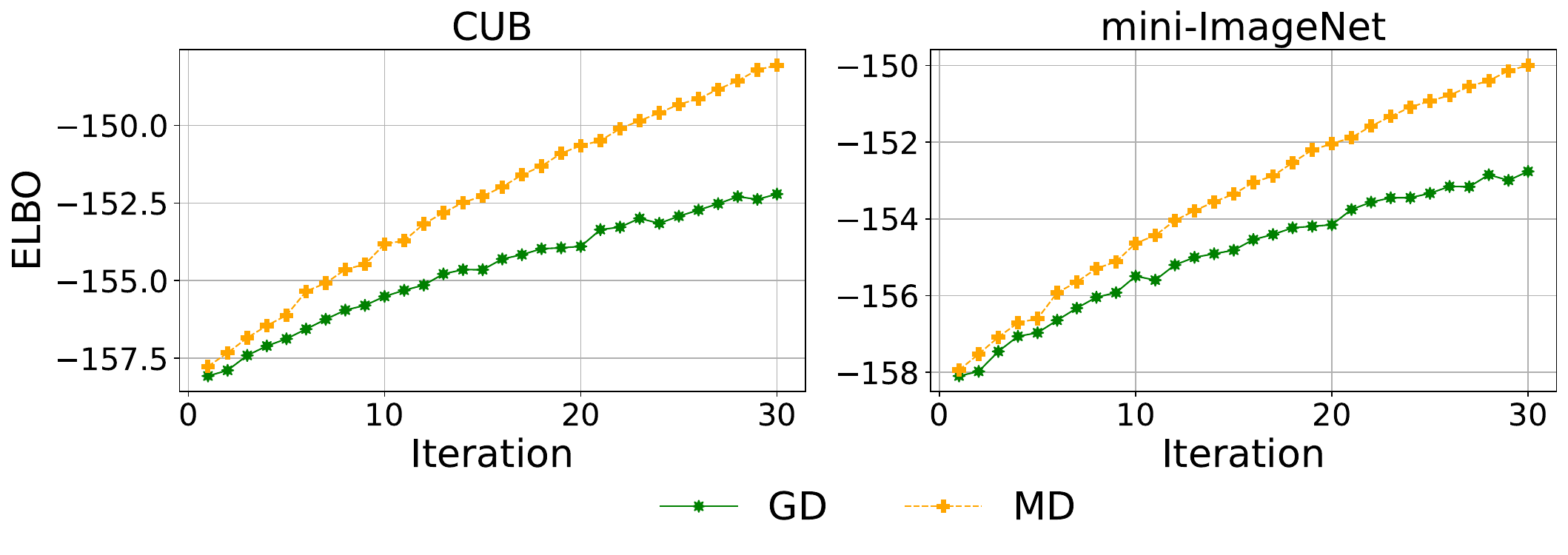}
        \label{fig:innerloop1shot}
    }
    \subfigure[5-shot classification]{
        \includegraphics[width = 0.48\textwidth]{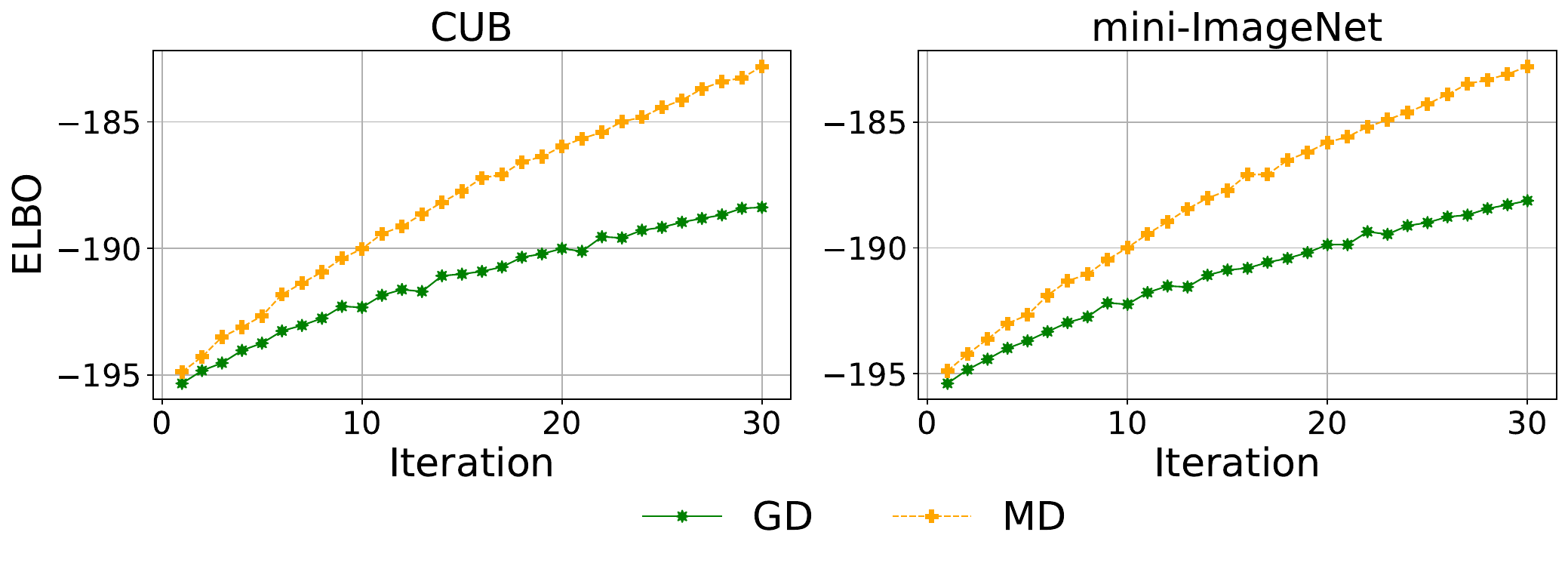}
        \label{fig:innerloop5shot}
    }
    \caption{The ELBO curve of 30 iterations on the CUB dataset and the mini-ImageNet dataset for 1-shot and 5-shot classification. Our method with mirror descent (MD) learns at a faster rate than the vanilla method with gradient descent (GD) in both scenarios.}
    \label{fig:innerloop}
\end{figure*}

\vspace{-5pt}
\paragraph{Inner-loop convergence}
For simplicity, we fix a random batch of data from a random episode for the CUB dataset and the mini-ImageNet dataset and observe the learning curve for ELBO. To ensure a fair comparison, we employ the same initialization scheme for the variational parameters and a comparable learning rate of $0.005$ for the inner loop and run each method for $30$ steps. Notice that in this empirical study, we employ a relatively small learning rate to better observe the alteration of the ELBO curve. The results can be seen in \cref{fig:innerloop}. 
The empirical observation aligns with our theoretical analysis: the ELBO of MD-BSFC increases at a faster rate than the vanilla method for both 1-shot and 5-shot scenarios. This is attributed to the fact that MD-BSFC employs second-order optimization in the inner loop, whereas the vanilla method utilizes first-order optimization. 

\begin{figure*}[t]
    \centering
    \subfigure[1-shot classification]{
        \includegraphics[width = 0.48\textwidth]{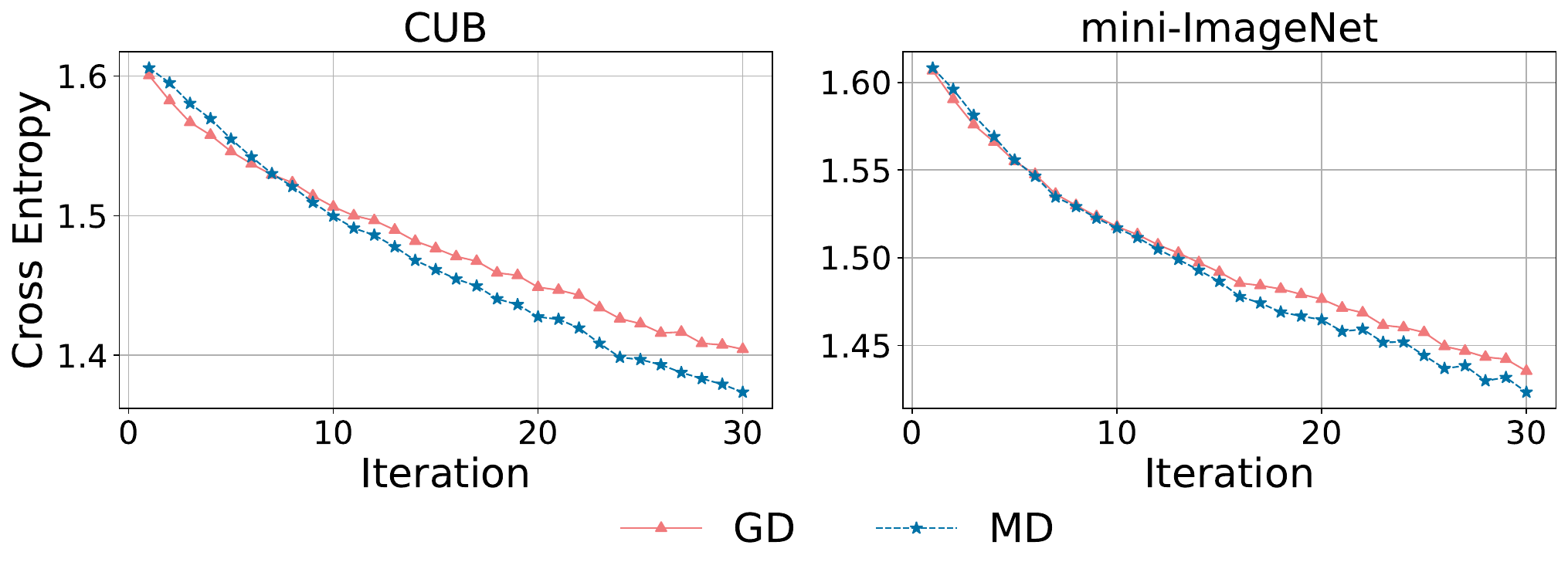}
        \label{fig:outerloop1shot}
    }
    \subfigure[5-shot classification]{
        \includegraphics[width = 0.48\textwidth]{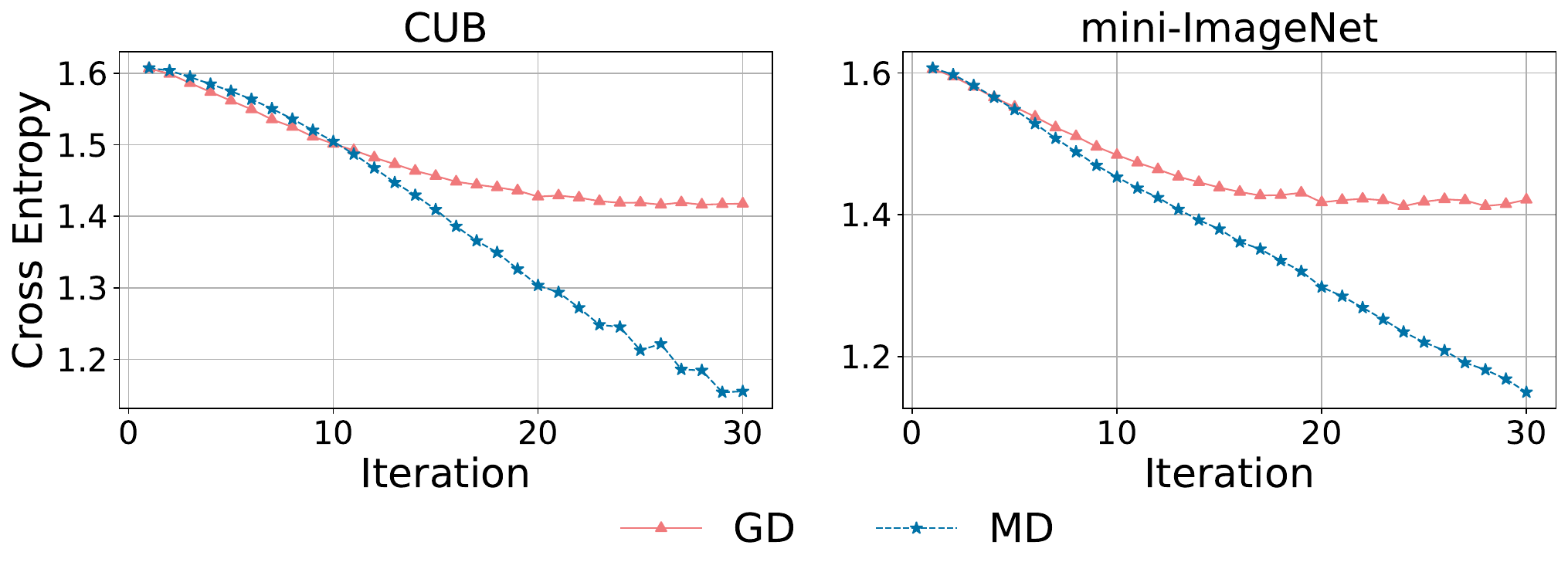}
        \label{fig:outerloop5shot}
    }
    \caption{The loss curve of 30 iterations on the CUB dataset and the mini-ImageNet dataset for 1-shot and 5-shot classification. Our method with mirror descent (MD) learns at a faster rate than the vanilla method with gradient descent (GD) in both scenarios.}
    \vspace{-8pt}
    \label{fig:outerloop}
\end{figure*}

\vspace{-5pt}
\paragraph{Outer-loop convergence}
For the full bi-level convergence experiment, we use a larger learning rate ($0.1$ for 1-shot and $0.02$ for 5-shot) with $2$ steps for the inner loop. The outer loop is updated by Adam with a learning rate of $0.001$. We empirically find that gradient descent displays high numerical instability for ELBO when a larger learning rate is applied. Thus, we use predictive likelihood for the outer-loop loss \citep{ke2023revisiting}, which is a plausible alternative to reflect the performance of convergence. Specifically, we perform inner-loop updates on the support samples and compute the cross-entropy loss on the validation samples. We run each method for $30$ iterations for the outer-loop updates as well and plot the cross-entropy loss. The results can be seen in \cref{fig:outerloop}. 
We observe that MD-BSFC takes a great lead in the 5-shot classification scenario and learns at a slightly higher pace in the 1-shot classification, demonstrating an improvement in convergence speed.

\subsection{Hyperparameter Settings}
In this section, we investigate the robustness of our method in different hyperparameter settings, including various inner-loop step size $\rho$, various kernels, and different number of steps for the inner loop. For simplicity, we present the results on the CUB dataset for the 1 shot scenario. 

%


\begin{table}
\centering
\caption{Accuracy of MD-BSFC in 1-shot 5-way tasks on the CUB dataset with different hyperparameters. Results are evaluated over 5 batches of 600 episodes with different random seeds.}
\scalebox{0.8}{
\begin{tabular}{cccccc}
     \toprule 
     $\bm{\rho}$ & \textbf{MD-BSFC} & \textbf{kernel} & \textbf{MD-BSFC} & \textbf{step} & \textbf{MD-BSFC} \\
    \midrule 
    0.01 & 32.22 $\pm$ 0.41 & COS & \textbf{65.45 $\pm$ 0.42} & 1 & 62.63 $\pm$ 0.26 \\
    0.1 & 61.54 $\pm$ 0.26 & RBF & 62.75 $\pm$ 0.39 & 3 & \textbf{65.45 $\pm$ 0.42} \\
    0.5 & 63.86 $\pm$ 0.38 & POL1 & 64.69 $\pm$ 0.36 & 5 & 64.18 $\pm$ 0.52  \\
    1 & \textbf{65.45 $\pm$ 0.42} & POL2 & 64.74 $\pm$ 0.25 & 7 & 62.55 $\pm$ 0.46 \\
    \bottomrule
\end{tabular}}
\label{tab:hyperparameter}
\end{table}
\vspace{-5pt}
\paragraph{Inner-loop step size}
Notice that in \cref{eq9}, the inner-loop step size $\rho$ controls the update rate of a single natural gradient step, where a larger $\rho$ may result in faster convergence but greater variance in general. However, in a bi-level optimization setting, the potential impact of varying step sizes combined with the highly nonconvex loss landscape of neural networks is not clear as opposed to single-level optimization. Therefore, we present some empirical results to shed some light on this issue in \cref{tab:hyperparameter}. Empirically, a larger $\rho$ leads to better performance, while for an extremely small $\rho$ the model tends to collapse. We suppose the phenomenon is due to a larger step size assist in escaping the saddle point of the highly nonconvex loss function.


\vspace{-5pt}
\paragraph{Base kernel}
\citet{massimi2020bayesian, jake2021bayesian, ke2023revisiting} illustrated the performance of various base kernels being close and the tendency of performance decay for smoother kernels, e.g., RBF kernel. We follow their protocol and test the results on similar candidate base kernels in \cref{tab:hyperparameter}, including cosine kernel with normalization (COS), RBF kernel, and polynomial kernel of order 1 (POL1) and 2 (POL2) \citep{massimi2020bayesian}. More details about the kernels are provided in \cref{sec:kernel}. Our empirical result conforms with the previous insights, where finite-rank kernels (POL1, POL2, COS) generally obtain a superior performance on the CUB dataset.
%

\vspace{-5pt}
\paragraph{Inner-loop steps}
We explore the inherent repercussions of varying inner-loop update steps. For single-level optimization where the outer-loop hyperparameters are fixed, the inner-loop parameters should be optimized until convergence. However, due to the complex training dynamics caused by bi-level optimization, it is not clear how we should select the inner-loop update steps. \citet{jake2017prototypical} employed a 1-step Gibbs sampling for training while \citet{ke2023revisiting} found a step number of 2 is optimal for various datasets for their method. Here we display some empirical results on this hyperparameter in \cref{tab:hyperparameter}.
Similar to \citet{ke2023revisiting}, we also find that a relatively small step number is optimal for our method. We suppose the reason behind this phenomenon is that only 1 step of update proffers an inferior approximation of the optimal task-specific parameters, while larger step numbers may affect the gradient flow. 

\vspace{-5pt}
\section{Limitation}

In this work, we focus on upgrading inner-loop optimization from a first-order to a second-order method, leading to enhanced overall convergence. However, the updates of the neural network (outer-loop) rely on a first-order method due to computational and memory constraints. Future work will explore second-order methods in outer-loop optimization. 


\vspace{-5pt}
\section{Conclusion}
In conclusion, our approach has contributed to Bayesian few-shot classification by addressing the non-conjugate inference challenge in GP classification. Moreover, the integration of non-Euclidean geometry through mirror descent has proven effective in enhancing the convergence rate theoretically and empirically. Experimental results robustly underscore the effectiveness of our proposed method, achieving competitive classification accuracy and uncertainty quantification on several benchmark datasets compared to baseline models. Additionally, our systematic exploration of various hyperparameters and components within the model provides valuable insights into their impact on performance. 


\section*{Impact Statement}
This paper presents work whose goal is to advance the field of Machine Learning. There are many potential societal consequences of our work, none of which we feel must be specifically highlighted here.

\section*{Acknowledgments}
This work was supported by NSFC Project (No. 62106121), the MOE Project of Key Research Institute of Humanities and Social Sciences (22JJD110001), and the Public Computing Cloud, Renmin University of China.

\bibliography{example_paper}
\bibliographystyle{icml2024}

\newpage
\appendix
\onecolumn
\section{Conjugate Variational Inference for Multi-class GP Classification}
\label{app1}
Consider a multi-class classification task consisting of $N$ observations with the input features $\mathbf{X}=[\mathbf{x}_1,\ldots,\mathbf{x}_N]^\top$ and the corresponding class labels $\mathbf{y}=[\mathbf{y}_1^\top,\ldots,\mathbf{y}_N^\top]^\top$, where $\mathbf{x}_i$ is a $D$-dimensional vector $\mathbf{x}_i\in\mathcal{X}\subset\mathbb{R}^D$ and $\mathbf{y}_i$ is the one-hot encoding for $C$ classes. 
The multi-class GP classification model includes latent GP functions for each class, i.e., $f^1,\ldots,f^C$ where $f^c(\cdot): \mathcal{X}\rightarrow\mathbb{R}$ is the corresponding latent function for $c$-th class. In order to address the multi-class classification with GPs, a GP prior is assumed for each latent function~\citep{williams2006gaussian}, i.e., $f^c\sim\mathcal{GP}(0,k_{\bm{\eta}_c})$ where $k_{\bm{\eta}_c}$ is the GP kernel for $c$-th class with hyperparameters $\bm{\eta}_c$. In practice, the hyperparameters are different for each class. 

Following tradition, we utilize the softmax link function to model the data likelihood (categorical distribution):
\begin{equation}
p(\mathbf{y}_n\vert\mathbf{f}_n)=\frac{\prod_{c}\exp{(f_n^c\cdot y_n^c)}}{\sum_{c}\exp{(f_n^c)}},
\label{app.eq1}
\end{equation}
where $f_n^c=f^c(\mathbf{x}_n)$ and $\mathbf{f}_n=[f_n^1,\ldots,f_n^C]^{\top}$. In order to predict the class label of a new data point $\mathbf{x}^*$,
we need to compute the posterior of latent functions using Bayesian framework:
\begin{equation}
p(\mathbf{f}\vert\mathbf{y})=\frac{p(\mathbf{y}\vert\mathbf{f})p(\mathbf{f})}{p(\mathbf{y})}=\frac{\prod_{n}p(\mathbf{y}_n\vert\mathbf{f}_n)\prod_{c}p(\mathbf{f}^c)}{\int\prod_{n}p(\mathbf{y}_n\vert\mathbf{f}_n)\prod_{c}p(\mathbf{f}^c)\dif\mathbf{f}},
\label{app.eq2}
\end{equation}
where $\mathbf{f}^c=[f_1^c,\ldots,f_N^c]^\top$, $\mathbf{f}=[\mathbf{f}^{1\top},\ldots,\mathbf{f}^{C\top}]^\top$, $p(\mathbf{f}^c)=\mathcal{N}(\mathbf{f}^c\vert\mathbf{0},\mathbf{K}^c)$ with $K_{ij}^c=k_{\bm{\eta}^c}(\mathbf{x}_i,\mathbf{x}_j)$. 

However, an issue is that the non-Gaussian likelihood in \cref{app.eq1} is non-conjugate to the Gaussian prior making the exact computation of posterior infeasible, so some approximate inference methods have to be utilized to approximate the posterior in \cref{app.eq2}. Variational inference (VI)~\citep{blei2017variational} is one of the most popular methods for approximate inference. VI can transform the inference problem into an optimization one which can be addressed by optimization techniques. Specifically, in VI, \cref{app.eq2} is approximated by a Gaussian distribution $q$:
\begin{equation*}
q(\mathbf{f}\vert\bm{\theta})=\prod_c\mathcal{N}(\mathbf{f}^c\vert\bm{\theta}^c),
\label{app.eq3}
\end{equation*}
where $\bm{\theta}=\{\bm{\theta}^c\}_{c=1}^C$ is the natural parameter for each class. This posterior approximation assumes independence among different latent functions of the model. The optimal Gaussian distribution is obtained by minimising the Kullback-Leibler (KL) divergence between $q$ and the exact posterior or equivalently maximizing the evidence lower bound (ELBO)~\citep{bishop2006pattern}:
\begin{equation}
\begin{aligned}
\argmax_{\bm{\theta}\in\bm{\Theta}}\mathcal{L}(\bm{\theta})&\coloneqq\mathbb{E}_{q}\left[\log p(\mathbf{y}\vert\mathbf{f})p(\mathbf{f})-\log q(\mathbf{f})\right]\\
&=\sum_{n}\mathbb{E}_{q}[\log p(\mathbf{y}_n\vert\mathbf{ f}_n)]-\sum_{c}\text{KL}[\mathcal{N}(\mathbf{f}^c\vert\bm{\theta}^c)\Vert\mathcal{N}(\mathbf{f}^c\vert\mathbf{0},\mathbf{K}^c)],
\end{aligned}
\label{app.eq4}
\end{equation}
where $\bm{\Theta}$ is the set of valid variational parameters and $\text{KL}$ is the KL divergence. 

One problem that arises when computing the ELBO in \cref{app.eq4} is that the expectation of log-likelihood, i.e., $\mathbb{E}_{q}[\log p(\mathbf{y}_n\vert\mathbf{ f}_n)]$, does not have a closed-form expression. Despite its intractability, the ELBO can still be optimized
by Monte Carlo combined with the reparametrization trick~\citep{diedrik2014auto}, but the major issues are: \textbf{(1)} A naive implementation of numerical optimization incurs low efficiency due to the complicated and time-consuming non-conjugate computation. \textbf{(2)} The rate of convergence depends on the parameterization of the variational distribution. The ordinary gradient in the Euclidean space is an unnatural direction to follow for VI because the parameters correspond to distributions and our goal is to optimize a distribution rather than its parameters~\citep{hoffman2013stochastic,salimbeni2018natural}. 

Inspired by~\citet{khan2017conjugate}, to address these two issues mentioned above, we propose an algorithm to optimize the ELBO with mirror descent developed by~\citet{nemirovskij1983problem} that combines the best of both worlds: on the one hand, each gradient step in our method
can be implemented in a conjugate way by approximating the non-conjugate term by a Gaussian distribution; on the other hand, our approach can maximize the ELBO more efficiently by exploiting the non-Euclidean geometry as the mirror descent is the steepest descent direction
along the corresponding Riemannian manifold and invariant to the parameterization of the variational distribution~\citep{raskutti2015information}.

\section{Derivation of Theorem 3.1}
\label{app2}
Consider a exponential-family distribution $p(\mathbf{f}\vert\bm{\theta})=h(\mathbf{f})\exp(\bm{\theta}^\top \phi(\mathbf{f})-A(\bm{\theta}))$ where $\bm{\theta}$ is the natural parameter,
$\phi(\mathbf{f})$ is the sufficient statistics, $A(\bm{\theta})$ is the log-partition function $A(\bm{\theta})=\log\int h(\mathbf{f})\exp(\bm{\theta}^\top \phi(\mathbf{f}))d\mathbf{f}$ that is convex. A minimal exponential family distribution can also be parameterized by the mean parameter $\bm{\mu}$ which is the mean of the sufficient statistics $\bm{\mu}=\mathbb{E}_p[\phi(\mathbf{f})]$. Considering the negative entropy of the distribution as a function of the mean parameter $H(\bm{\mu})=\mathbb{E}_p[\log p(\mathbf{f}\vert\bm{\theta})]=\bm{\theta}^\top\bm{\mu}-A(\bm{\theta})$ where we omit the constant term, it is a well known result that the log partition function and negative entropy function are convex conjugate functions: $H(\bm{\mu})=\sup_{\bm{\theta}^\prime}[\bm{\theta}^{\prime\top}\bm{\mu}-A(\bm{\theta}^\prime)]$ and $A(\bm{\theta})=\sup_{\bm{\mu}^\prime}[\bm{\theta}^{\top}\bm{\mu}^\prime-H(\bm{\mu}^\prime)]$. Further, if $A(\cdot)$ is strictly convex and twice differentiable, then so is $H(\cdot)$. These two convex conjugate
functions provide an important relation between natural parameter and mean parameter: $\bm{\mu}=\nabla A(\bm{\theta})$ and $\bm{\theta}=\nabla H(\bm{\mu})$~\citep{rockafellar2015convex}.

The two convex conjugate functions will induce a pair of dual Bregman divergences, and then the dual Bregman divergences will further induce a pair of dual Riemannian manifolds. Given the log-partition function $A(\bm{\theta})$, it induces a Bregman divergence $B_A(\bm{\theta},\bm{\theta}^\prime):\bm{\Theta}\times\bm{\Theta}\rightarrow\mathbb{R}^+$ that defines a positive definite Riemannian metric $\nabla^2A(\bm{\theta})$ because $A(\bm{\theta})$ is a strictly convex and twice differentiable function. Therefore, $B_A(\bm{\theta},\bm{\theta}^\prime)$ induces the Riemannian manifold $(\bm{\Theta},\nabla^2A(\bm{\theta}))$. Similarly, if all $\bm{\theta}\in\bm{\Theta}$ are transformed by $\nabla A(\bm{\theta})$, we can get the mean parameter $\bm{\mu}\in\mathcal{M}$. The Bregman divergence $B_H(\bm{\mu},\bm{\mu}^\prime):\mathcal{M}\times\mathcal{M}\rightarrow\mathbb{R}^+$ induces a Riemannian manifold $(\mathcal{M},\nabla^2H(\bm{\mu}))$. $(\bm{\Theta},\nabla^2A(\bm{\theta}))$ is called the primal Riemannian manifold and $(\mathcal{M},\nabla^2H(\bm{\mu}))$ is called the dual Riemannian manifold. \citet{raskutti2015information} proved the equivalence between mirror descent and natural gradient descent by the application of chain rule and dual Riemannian manifolds. We restate the proof here. 
\begin{proof}
Assume that we use a mirror descent to maximize the ELBO $\tilde{\mathcal{L}}(\bm{\mu})=\mathcal{L}(\bm{\theta})$ over the mean parameter $\bm{\mu}$ using the Bregman divergence induced by the negative entropy function $H(\bm{\mu})$:
\begin{equation*}
\begin{aligned}
\bm{\mu}_{t+1}=\argmax_{\bm{\mu}\in\mathcal{M}}\nabla \tilde{\mathcal{L}}(\bm{\mu}_t)^\top\bm{\mu}-\frac{1}{\rho_t}B_{H}(\bm{\mu},\bm{\mu}_t). 
\end{aligned}
\end{equation*}
Substituting $B_{H}(\bm{\mu},\bm{\mu}_t)=H(\bm{\mu})-H(\bm{\mu}_t)-\nabla H(\bm{\mu}_t)^\top(\bm{\mu}-\bm{\mu}_t)$, we can obtain the solution:
\begin{equation*}
\begin{aligned}
\nabla H(\bm{\mu}_{t+1})=\nabla H(\bm{\mu}_{t})+\rho_t\nabla \tilde{\mathcal{L}}(\bm{\mu}_t).
\end{aligned}
\end{equation*}
Using $\bm{\mu}=\nabla A(\bm{\theta})$ and $\bm{\theta}=\nabla H(\bm{\mu})$, we obtain the following equivalent expression:
\begin{equation*}
\begin{aligned}
\bm{\theta}_{t+1}=\bm{\theta}_{t}+\rho_t\nabla_{\bm{\mu}} \tilde{\mathcal{L}}(\nabla A(\bm{\theta}_t)).
\end{aligned}
\end{equation*}
Substituting the chain rule $\nabla_{\bm{\theta}}\tilde{\mathcal{L}}(\nabla A(\bm{\theta}))=\nabla^2_{\bm{\theta}}A(\bm{\theta})\nabla_{\bm{\mu}}\tilde{\mathcal{L}}(\nabla A(\bm{\theta}))$, we obtain:
\begin{equation*}
\begin{aligned}
\bm{\theta}_{t+1}=\bm{\theta}_{t}+\rho_t[\nabla^2_{\bm{\theta}}A(\bm{\theta}_t)]^{-1}\nabla_{\bm{\theta}}\mathcal{L}(\bm{\theta}_t), 
\end{aligned}
\end{equation*}
which is just the natural gradient descent in \cref{eq4}. 
\end{proof}

\section{Gradients w.r.t. Mean Parameters}
\label{app3}
In this section, we provide a proof of \cref{eq10,eq11} in the main paper.
\begin{proof}
Given the softmax likelihood: $p(\mathbf{y}_n\vert\mathbf{f}_n)=\prod_{c}\exp{(f_n^c\cdot y_n^c)}/\sum_{c}\exp{(f_n^c)}$, we can obtain the log-likelihood and the first and the diagonal of second-order gradients of log-likelihood w.r.t. $\mathbf{f}_n$:
\begin{gather*}
\log p(\mathbf{y}_n\vert\mathbf{f}_n)=\mathbf{f}_n\cdot\mathbf{y}_n-\log(\sum_c\exp(\mathbf{f}_n)),\\
\nabla_{\mathbf{f}_n}\log p(\mathbf{y}_n\vert\mathbf{f}_n)=\mathbf{y}_n-\frac{\exp(\mathbf{f}_n)}{\sum_c\exp(\mathbf{f}_n)},\\
\text{diag}(\nabla_{\mathbf{f}_n}^2\log p(\mathbf{y}_n\vert\mathbf{f}_n))=\frac{\exp(2\mathbf{f}_n)}{(\sum_c\exp(\mathbf{f}_n))^2}-\frac{\exp(\mathbf{f}_n)}{\sum_c\exp(\mathbf{f}_n)}.
\end{gather*}
Given a factorized variaitonal distribution $q(\mathbf{f}\vert\bm{\theta})=\prod_c\mathcal{N}(\mathbf{f}^c\vert\bm{\theta}^c)$, the ELBO can be written as:
\begin{equation*}
\begin{aligned}
\mathcal{L}=\sum_n\mathbb{E}_q[\log p(\mathbf{y}_n\vert\mathbf{f}_n)]-\sum_c\text{KL}[\mathcal{N}(\mathbf{f}^c\vert\bm{\theta}^c)\Vert\mathcal{N}(\mathbf{f}^c\vert\mathbf{0},\mathbf{K}^c)].
\end{aligned}
\end{equation*}
If we parameterize $q(\mathbf{f}_n)$ as $\mathcal{N}(\mathbf{f}_n\vert\mathbf{m}_n,\text{diag}(\mathbf{v}_n))$, where $\text{diag}(\mathbf{v}_n)$ is a diagonal matrix corresponding to vector $\mathbf{v}_n$ due to the assumption of independence between latent functions of different classes, the gradient of $\mathbb{E}_{q(\mathbf{f}_n)}[\log p(\mathbf{y}_n\vert\mathbf{f}_n)]$ w.r.t. $\mathbf{m}_n$ and $\mathbf{v}_n$ can be expressed as: 
\begin{gather*}
\mathbf{g}_{\mathbf{m}_n}\coloneqq\nabla_{\mathbf{m}_n}\mathbb{E}_{q(\mathbf{f}_n)}[\log p(\mathbf{y}_n\vert\mathbf{f}_n)]=\mathbb{E}_q\nabla_{\mathbf{f}_n}\log p(\mathbf{y}_n\vert\mathbf{f}_n)=\mathbb{E}_q\left[\mathbf{y}_n-\frac{\exp{(\mathbf{f}_n)}}{\sum_c \exp{(\mathbf{f}_n)}}\right],\\
\mathbf{g}_{\mathbf{v}_n}\coloneqq\nabla_{\mathbf{v}_n}\mathbb{E}_{q(\mathbf{f}_n)}[\log p(\mathbf{y}_n\vert\mathbf{f}_n)]=\frac{1}{2}\mathbb{E}_q\text{diag}(\nabla_{\mathbf{f}_n}^2\log p(\mathbf{y}_n\vert\mathbf{f}_n))=\frac{1}{2}\mathbb{E}_q\left[\frac{\exp{(2\mathbf{f}_n)}}{(\sum_c\exp{(\mathbf{f}_n)})^2}-\frac{\exp{(\mathbf{f}_n)}}{\sum_c\exp{(\mathbf{f}_n)}}\right].
\end{gather*}
according to \citet{opper2009variational}. 

Given the relation between the traditional parameters and the mean parameters of an independent multivariate Gaussian distribution: $\mathbf{m}_n=\bm{\mu}_n^{(1)}$ and $\mathbf{v}_n=\bm{\mu}_n^{(2)}-\bm{\mu}_n^{(1)2}$ where $\bm{\mu}_n^{(1)}$ and $\bm{\mu}_n^{(2)}$ are two mean parameters of $q(\mathbf{f}_n)$, we can use the chain rule to express the gradient w.r.t. the mean parameters:
\begin{gather*}
\nabla_{\bm{\mu}_n^{(1)}}\mathbb{E}_{q(\mathbf{f}_n)}[\log p(\mathbf{y}_n\vert\mathbf{f}_n)]=\mathbf{g}_{\mathbf{m}_n}-2\mathbf{g}_{\mathbf{v}_n}\circ\mathbf{m}_n,\\
\nabla_{\bm{\mu}_n^{(2)}}\mathbb{E}_{q(\mathbf{f}_n)}[\log p(\mathbf{y}_n\vert\mathbf{f}_n)]=\mathbf{g}_{\mathbf{v}_n}.
\end{gather*}
\end{proof}


\section{Computational Complexity for \cref{eq15a}}
Note that the computational complexity of Monte Carlo in \cref{eq15a} comes from covariance decomposition and sample drawing. The first part costs $O(MN^2)$ and the second part costs $O(N^3)$, where $N$ is the support size and $M$ is the samples drawn. In a few-shot setting, $M$ is much larger than $N$, so the total computational complexity is $O(MN^2)$, which does not pose a challenge to computational resources and is easily manageable.

\section{Experimental Details}
\label{sec:exp}
\subsection{Datasets}
We use four datasets as described below.
\begin{enumerate}
    \item \textbf{Caltech-UCSD Birds (CUB).} CUB consists of a total of 11788 bird images from 200 distinct classes. The standard split of 100 training, 50 validation, and 50 test classes is employed \citep{jake2021bayesian}.
    \item \textbf{mini-ImageNet.} mini-ImageNet is a small part of the full ImageNet dataset, where 100 classes with 600 images each are selected to form the dataset. We employed the common split of 64 training, 16 validation, and 20 test classes as well \citep{jake2021bayesian}.
    \item \textbf{Omniglot.} There are 1623 grayscale characters selected from 50 languages contained in this dataset, which is further expanded to 6492 images by data augmentation \citep{lake2011one}. We use 4114 for training.
    \item \textbf{EMNIST.} EMNIST \citep{cohen2017emnist} consists of 62 classes of single digits and English characters. In the domain transfer task, we utilize 31 for validation and the other for test.
\end{enumerate}

\subsection{Comparison of Baselines}
As for the description of baseline methods, we refer to \citet{jake2021bayesian} for a detailed overview. Here we only compare the methods that are most similar to ours, which include DKT, Logistic-softmax, and OVE. All of these methods utilize a similar framework but with different likelihood and inference methods from ours.
\begin{enumerate}
    \item \textbf{Deep Kernel Transfer (DKT)} \citep{massimi2020bayesian} employed the Gaussian likelihood to circumvent the conjugacy issue where the labels $\{+1, -1\}$ are treated as continuous values, leading to suboptimal accuracy.
    \item \textbf{Logistic-softmax} \citep{galy2020multi} applied the logistic-softmax for a conditional conjugate model after data augmentation. The Gibbs sampling version implemented by \citet{jake2021bayesian} and the mean-field approximation version implemented by \citet{ke2023revisiting} are considered for comparison.
    \item \textbf{One-vs-Each Approximation (OVE)} \citep{jake2021bayesian} proposed to approximate the softmax function for conditional conjugacy after data augmentation and applied Gibbs sampling for posterior inference. 
\end{enumerate}

\subsection{Training Protocols}
The Adam optimizer with a standard learning rate of $10^{-3}$ for the neural network and a learning rate of $10^{-4}$ for other kernel parameters is employed across all our experiments in the outer loop. For a single epoch, 100 random episodes are sampled from the complete dataset for all methods. As for the steps used for variational inference, we run 3 steps with $\rho = 1$ during training time and 50 steps during testing time with $\rho = 0.5$. We employ the COS kernel for CUB and mini-ImageNet and the RBF kernel for Omniglot and EMNIST.

\subsection{Kernels}
\label{sec:kernel}

To begin with, we briefly review the definition of deep kernel proposed in \citet{wilson2016deep}. Deep kernel combines the traditional kernel with modern neural networks in a straightforward formulation, defined as $k(\mathbf{x},\mathbf{x}'\mid\bm{\theta},\mathbf{w})=k'(g_{\mathbf{w}}(\mathbf{x}),g_{\mathbf{w}}(\mathbf{x}')\mid\bm{\theta})$, where $k'$ represents a predefined base kernel with parameters $\bm{\theta}$. The inputs are projected by a deep neural network to increase flexibility. Note that the deep kernel parameters consist of $\bm{\theta}$ and $\mathbf{w}$. One upside of deep kernel is its potential to learn beyond Euclidean distance-based metrics via data-driven optimization. 
In the following, we provide the expressions for the relevant base kernels discussed in the main paper. 

\begin{enumerate}
    \item \textbf{Polynomial Kernel (POL).} The polynomial kernel is defined as
        $$
        k^{\prime}(\mathbf{x}, \mathbf{x}^{\prime})=(\mathbf{x}^{\top} \mathbf{x}^{\prime}+c)^s,
        $$
    where $s$ is the order of the kernel and $c$ is a trainable parameter. 
    \item \textbf{Radial Basis Function Kernel (RBF).} The RBF kernel is defined as
        $$
        k^{\prime}(\mathbf{x}, \mathbf{x}^{\prime})=\exp \Big(-\frac{\left\|\mathbf{x}-\mathbf{x}^{\prime}\right\|^2}{2 l^2}\Big),
        $$
    where $l$ is referred to as the length-scale parameter. The RBF kernel is smooth because of the exponential eigenvalue decay rate. 
    \item \textbf{Cosine Similarity Kernel (COS).} The cosine similarity kernel is defined as 
    $$
        k^{\prime}(\mathbf{x}, \mathbf{x}^{\prime})=\frac{\mathbf{x}^\top \mathbf{x}^{\prime}}{\|\mathbf{x}\|\left\|\mathbf{x}^{\prime}\right\|}.
    $$
    We use a variant of the kernel with batch normalization that centers the input vectors \citep{massimi2020bayesian}.
\end{enumerate}

\subsection{Comparision of Actual Running Time}
To further investigate the convergence rate in practice, we give a brief comparison of the actual running time between our method and other first-order methods in \cref{tab:runningtime}. We use the CUB dataset and run each method with 3 steps for the inner loop except CDKT (PL). The time reported for the inner loop and outer loop are from the last sampled task for the first epoch. We use one Quadro RTX 6000 to run each method. Note that CDKT (ML) is one of the sota baselines that use a very similar training procedure, where its inner loop is the mean-field approximation steps, and CDKT (PL) uses predictive likelihood with no inner loop so we just compare the total run time of an epoch. As can be seen from the table, our method has a similar run time compared to other first-order method. The result proves that our method indeed enjoys first-order computation and is comparable to other first-order methods in actual running time.

\begin{table}
\centering
\caption{Comparison of actual running time between MD-BSFC and other first-order methods with similar implementations on the CUB dataset. The inner loop is run with 3 steps.}
\scalebox{0.85}{
\begin{tabular}{ccccc}
     \toprule 
     & \textbf{MD} & \textbf{GD} & \textbf{CDKT (ML)} & \textbf{CDKT(PL)} \\
    \midrule 
    Inner Loop & 0.0181 & 0.0172 & 0.0141 & NA \\
    Outer Loop & 0.1035	& 0.0989 & 0.0814 & NA \\
    One Epoch&12.1253&12.0518&10.8549&12.5445 \\
    \bottomrule
\end{tabular}}
\label{tab:runningtime}
\end{table}

\section{Related Work}
To address the non-conjugate issue, \citet{massimi2020bayesian} assumes a Gaussian likelihood for class labels to circumvent the non-conjugate computation, which exhibits inferior uncertainty quantification since class labels are discrete rather than continuous. Differently, \citet{jake2021bayesian} proposes a novel approach that combines the P\'{o}lya-Gamma augmentation technique~\citep{polson2013bayesian} and the one-vs-each softmax approximation~\citep{titsias2016onevseach} to transform the non-conjugate GP classification model into a conditionally conjugate model. \citet{ke2023revisiting} also utilizes the P\'{o}lya-Gamma augmentation with the logistic-softmax likelihood to achieve conditional conjugacy. These methods offer better uncertainty quantification but require additional auxiliary variables. 
Differing from the approaches mentioned above, we incorporate mirror descent-based variational inference into GP classification to obtain a conjugate algorithm without introducing any auxiliary variables. 

The exploration of Bayesian meta-learning has yielded diverse approaches aimed at leveraging prior knowledge and adapting to new tasks. The framework proposed by \citet{finn2018probabilistic} adopts a Bayesian hierarchical modeling perspective, enabling the capture of uncertainty at various levels. In a different vein, \citet{grant2018recasting} reformulated meta-learning as inference in a GP, while \citet{yoon2018bayesian} introduced Bayesian Model-Agnostic Meta-Learning (MAML) building upon the work of \citet{finn2017model}. Notably, \citet{massimi2020bayesian,jake2021bayesian} employed GPs with deep kernels to facilitate task-specific inference, thereby contributing to Bayesian meta-learning in terms of parameter updates, uncertainty modeling, and prior distributions. In this context, our work makes a valuable contribution by presenting an effective alternative for task-level updates and offering insights into the coordination problem associated with bi-level optimization.

\end{document}